\documentclass{article}
\usepackage{iclr2023_conference,times}
\iclrfinalcopy
\usepackage[utf8]{inputenc} 
\usepackage[T1]{fontenc}    
\usepackage[colorlinks=true,linkcolor=blue,citecolor=blue,urlcolor=blue]{hyperref} 
\usepackage{url}            
\usepackage{booktabs}       
\usepackage{amsfonts}       
\usepackage{nicefrac}       
\usepackage{microtype}      
\usepackage{xcolor}         

\usepackage{amsmath}
\usepackage{amssymb}
\usepackage{amsthm}
\usepackage{multirow}
\usepackage{array}
\usepackage{graphics}
\usepackage{mathtools}
\usepackage{enumerate}
\usepackage{xspace}
\usepackage{wrapfig}
\usepackage{pifont}
\usepackage{float}
\usepackage{nicefrac}
\usepackage{bm}
\usepackage{bbm}
\usepackage{algorithm}
\usepackage{algorithmic}
\usepackage{subfigure}

\usepackage[toc,page,header]{appendix} 
\usepackage{minitoc}

\definecolor{red}{RGB}{215,48,39}
\definecolor{green}{RGB}{26,152,80}
\definecolor{lightgray}{gray}{0.96}
\definecolor{blue}{RGB}{30, 144, 255}
\newtheorem{theorem}{Theorem}
\newtheorem{lemma}[theorem]{Lemma}

\DeclareMathOperator*{\argmin}{arg\,min}
\newcommand*{\defeq}{\stackrel{\text{def}}{=}}

\theoremstyle{definition}

\title{Quantifying and Mitigating the Impact of Label Errors on Model Disparity Metrics}

\author{%
  Julius Adebayo\\ Prescient Design / Genentech \And 
  Melissa Hall \\ Meta \And
  Bowen Yu \\ Meta \And
  Bobbie Chern \\ Meta
}

\begin{document}

\doparttoc 
\faketableofcontents 

\maketitle

\begin{abstract}
Errors in labels obtained via human annotation adversely affect a model's performance. Existing approaches propose ways to mitigate the effect of label error on a model's downstream accuracy, yet little is known about its impact on a model's disparity metrics\footnote{Group-based disparity metrics like subgroup calibration, false positive rate, false negative rate, equalized odds, and equal opportunity are more often known, colloquially, as \textit{fairness metrics} in the literature. We use the term group-based disparity metrics in this work.}. Here we study the effect of label error on a model's disparity metrics. We empirically characterize how varying levels of label error, in both training and test data, affect these disparity metrics. We find that group calibration and other metrics are sensitive to train-time and test-time label error---particularly for minority groups. This disparate effect persists even for models trained with noise-aware algorithms. To mitigate the impact of training-time label error, we present an approach to estimate the \textit{influence} of a training input's label on a model's group disparity metric. We empirically assess the proposed approach on a variety of datasets and find significant improvement, compared to alternative approaches, in identifying training inputs that improve a model's disparity metric. We complement the approach with an automatic relabel-and-finetune scheme that produces updated models with, provably, improved group calibration error.
\end{abstract}

\section{Introduction}
\label{introduction}
Label error (noise) --- mistakes associated with the label assigned to a data point --- is a pervasive problem in machine learning~\citep{northcutt2021pervasive}. For example, 30 percent of a random 1000 samples from the Google Emotions dataset~\citep{demszky2020goemotions} had label errors~\citep{chen2022emotions}. Similarly, an analysis of the MS COCO dataset found that up to 37 percent (273,834 errors) of all annotations are erroneous~\citep{murdoch2022cocodataset}. Yet, little is known about the effect of label error on a model's group-based disparity metrics like equal odds~\citep{hardt2016equality}, group calibration~\citep{pleiss2017fairness}, and false positive rate~\citep{barocas-hardt-narayanan}. 

It is now common practice to conduct `fairness' audits (see:~\citep{buolamwini2018gender, raji2019actionable, bakalar2021fairness}) of a model's predictions to identify data subgroups where the model underperforms. Label error in the test data used to conduct a fairness audit renders the results unreliable. Similarly, label error in the training data, especially if the error is systematically more prevalent in certain groups, can lead to models that associate erroneous labels to such groups. The reliability of a fairness audit rests on the assumption that labels are \textit{accurate}; yet, the sensitivity of a model's disparity metrics to label error is still poorly understood. Towards such end, we ask:\\

\centerline{\textit{what is the effect of label error on a model's disparity metric?}}

We address the high-level question in a two-pronged manner via the following questions:
\begin{enumerate}
    \item \textbf{Research Question 1}: What is the sensitivity of a model's disparity metric to label errors in training and test data? Does the effect of label error vary based on group size?
    \item \textbf{Research Question 2}: How can a practitioner identify training points whose labels have the most \textit{influence} on a model's group disparity metric?
\end{enumerate}

\subsection*{Contributions \& Summary of Findings}
\begin{figure*}[th]
\centering
\includegraphics[scale=0.35]{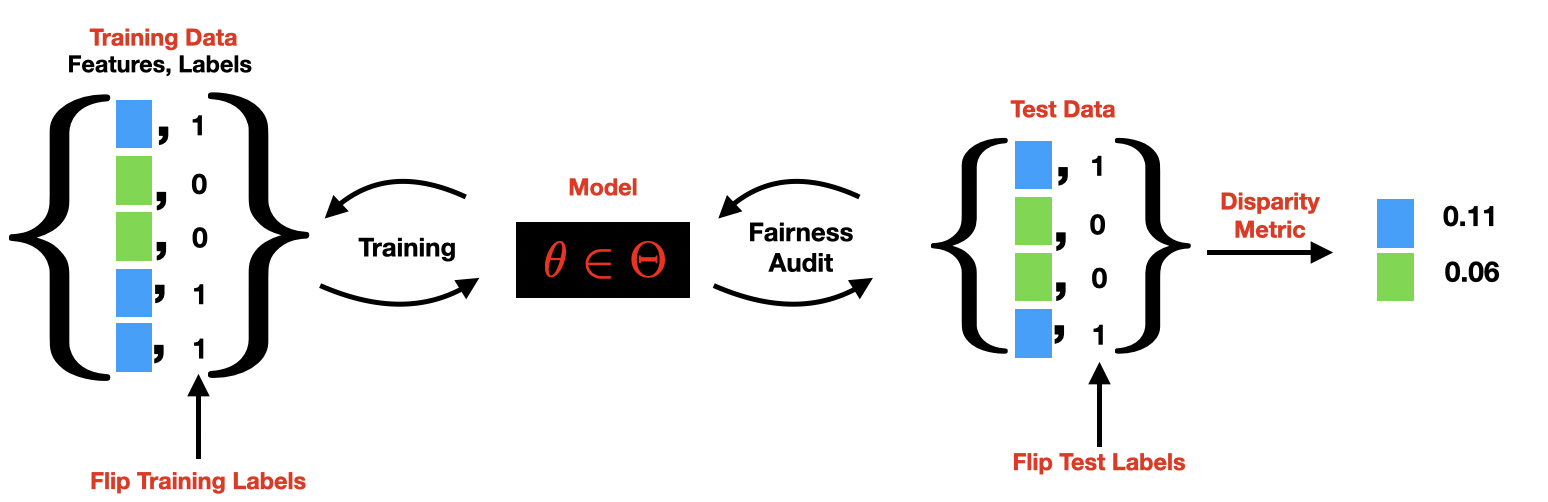}
\caption{\textbf{A schematic of the test and train-time empirical sensitivity tests}. Here we show the model training and fairness audit pipeline. Our proposed sensitivity tests capture the effect of label error, in both stages, on the disparity metric. In the Test-time sensitivity test, we flip the label of a portion of the test data and then compare the corresponding disparity metric (group calibration for example) for the flipped dataset to the metrics for a standard model where the test labels were not flipped. In the Train-time sensitivity test, we flip the labels of a portion of the training set, and then measure the change in disparity metric to a standard model.}
\label{fig:schematic}
\end{figure*}
In addressing these questions, we make two broad contributions: 

\textbf{Empirical Sensitivity Tests.} We assess the sensitivity of model disparity metrics to label errors with a label flipping experiment. First, we iteratively flip the labels of samples in the test set, for a fixed model, and then measure the corresponding change in the model disparity metric compared to an unflipped test set. Second, we fix the test set for the fairness audit but flip the labels of a proportion of the training samples. We then measure the change in the model disparity metrics for a model trained on the data with flipped labels. We perform these tests across a datasets and model combinations. 

\textbf{Training Point Influence on Disparity Metric.} We propose an approach, based on a modification to the influence of a training example on a test example's loss, to identify training points whose labels have undue effects on any disparity metric of interest on the test set. We empirically assess the proposed approach on a variety of datasets and find a 10-40\% improvement, compared to alternative approaches that focus solely on model's loss, in identifying training inputs that improve a model's disparity metric.

\section{Setup \& Background}
\label{background}
In this section, we discuss notation, and set the stage for our contributions by discussing the disparity metrics that we focus on. We also provide an overview of the datasets and models used in the experimental portions of the paper.\footnote{We refer readers to the longer version of this work on the arxiv. Code to replicate our findings is available at: \url{https://github.com/adebayoj/influencedisparity}}

\textbf{Overview of Notation.} We consider prediction problems, i.e, settings where the task is to learn a mapping, $\theta: \mathcal{X} \times \mathcal{A} \rightarrow \mathcal{Y}$, where $\mathcal{X} \in \mathbb{R}^d$ is the feature space, $\mathcal{Y} \in \{0, 1\}$ is the output space, and $\mathcal{A}$  is a group identifier that partitions the population into disjoint sets e.g. race, gender. We can represent the tuple $(x_i, a_i, y_i)$ as $z_i$. Consequently, the $n$ training points can be written as: $\{z_i\}_{i=1}^{n}$. Throughout this work, we will only consider learning via empirical risk minimization (ERM), which corresponds to: $\hat{\theta} := \argmin_{\theta \in \Theta} \frac{1}{n}\sum_{i}^{n}\ell(z_i, \theta)$. Similar to \citet{koh2017understanding}, we will assume that the ERM objective is twice-differentiable and strictly convex in the parameters. We focus on binary classification tasks, however, our analysis can be easily generalized.

\textbf{Disparity Metrics.} We define a group disparity metric to be a function, $\mathcal{GD}$, that gives a performance score given a model's probabilistic predictions ($\theta$ outputs the probability of belonging to the positive class) and `ground-truth' labels. We consider the following metrics (We refer readers to the Appendix for a detailed overview of these metrics):

\begin{enumerate}
    \item \textbf{Calibration}: defined as $\mathbb{P}\left( \hat{y}=y \vert \hat{p} = p \right), \forall p \in [0, 1]$. In this work, we measure calibration with two different metrics: 1) Expected Calibration Error (ECE)~\citep{naeini2015obtaining, pleiss2017fairness}, and 2) the Brier Score~\citep{rufibach2010use} (BS).
    \item \textbf{ (\textit{Generalized}) False Positive Rate (FPR)}:   is $\mathcal{GD}_{\mathrm{fpr}}(\theta) = \mathbb{E}[\theta(x_i)~\vert~y_i = 0]$ (see~\cite{guo2017calibration}),
    \item \textbf{ (\textit{Generalized}) False Negative Rate (FNR)}: is $\mathcal{GD}_{\mathrm{fnr}}(\theta) = \mathbb{E}[(1 - \theta(x_i))~\vert~y_i = 1]$,
    \item \textbf{Error Rate (ER)}: is the $\mathcal{GD}_{\mathrm{er}}(\theta) = 1 - \mathrm{acc}(\theta)$.
\end{enumerate}

We consider these metrics separately for each group as opposed to relative differences. For each dataset, we consider the protected data subgroup with the largest size as the majority group, and the group  the smallest size is the minority group.

\textbf{Datasets.} We consider datasets across different modalities: 4 tabular, and a text dataset. A description of these datasets along with test accuracy is provided in Table~\ref{tab:data-characteristics}. Each dataset contains annotations with a group label for both training and test data, so we are able to manipulate these labels for our empirical sensitivity tests. For the purposes of this work, we assume that the provided labels are the ground-truth---a strong assumption that nevertheless does not impact the interpretation of our findings.
\begin{table*}[t]
\centering
\begin{tabular}{lrrrrl}
Dataset          & Classes & $n$       & $d$     & Group & Source \\
\hline
CivilComments        &  $2$    & $1,820,000$  & $768$   & Sex   & \citet{koh2017understanding} \\
ACSIncome            &  $2$    & $195,665$   & $10$ & Sex, Race   & \citet{ding2021retiring} \\
ACSEmployment         &  $2$    & $378,817$   & $16$ & Sex, Race   & \citet{ding2021retiring} \\
ACSPublic Coverage           &  $2$   & $138,554$  & $19$   & Sex, Race  & \citet{ding2021retiring} \\
Credit Dataset              &  $2$    & $405,032$  & $6$   & Sex  & \citet{de2015unique}
\end{tabular}
\caption{Overview of dataset characteristics for the datasets considered in this work.}
\label{tab:data-characteristics}
\end{table*}

\textbf{Model.} We consider three kinds of model classes in this work: 1) a logistic regression model, 2) a Gradient-boosted Tree (GBT) classifier for the tabular datasets, and 3) a ResNet-18 model. We only consider the logistic regression and GBT models for tabular data, while we fine-tune a ResNet-18 model on embeddings for the text data.

\section{Empirical Assessment of Label Sensitivity}
\label{empirical}
In this section, we perform empirical sensitivity tests to quantify the impact of label error on test group disparity metrics. We conduct tests on data from two different stages of the ML pipeline: 1) Test-time (test dataset) and 2) Training-time (training data). We use as our primary experimental tool: label flipping, i.e., we flip the labels of a percentage of the samples, uniformly at random in either the test or training set, and then measure the concomitant change in the model disparity metric. We assume that each dataset's labels are the ground truth and that flipping the label results in label error for the samples whose labels have been overturned. Recent literature has termed this setting synthetic noise, i.e., the label flipping simulates noise that might not be representative of real-world noise in labels~\citep{arpit2017closer, zhang2021understanding, jiang2020beyond}. 

\subsection{Sensitivity to Test-time Label Error}
\textbf{Overview \& Experimental Setup.} The goal of the test-time empirical test is to measure the impact of label error on the group calibration error of a fixed model. Consider the setting where a model has been trained, and a fairness assessment is to be conducted on the model. What impact does label error, in the test set used to conduct the audit, have on the calibration error on the test data? The test-time empirical tests answer this question. Given a fixed model, we iteratively flip a percentage of the labels, uniformly at random, ranging from zero to 30 percent in the test data. We then estimate the model’s calibration using the modified dataset. Critically, we keep the model fixed while performing these tests across each dataset. 

\begin{figure*}[!h]
\centering
\includegraphics[scale=0.25, page=3]{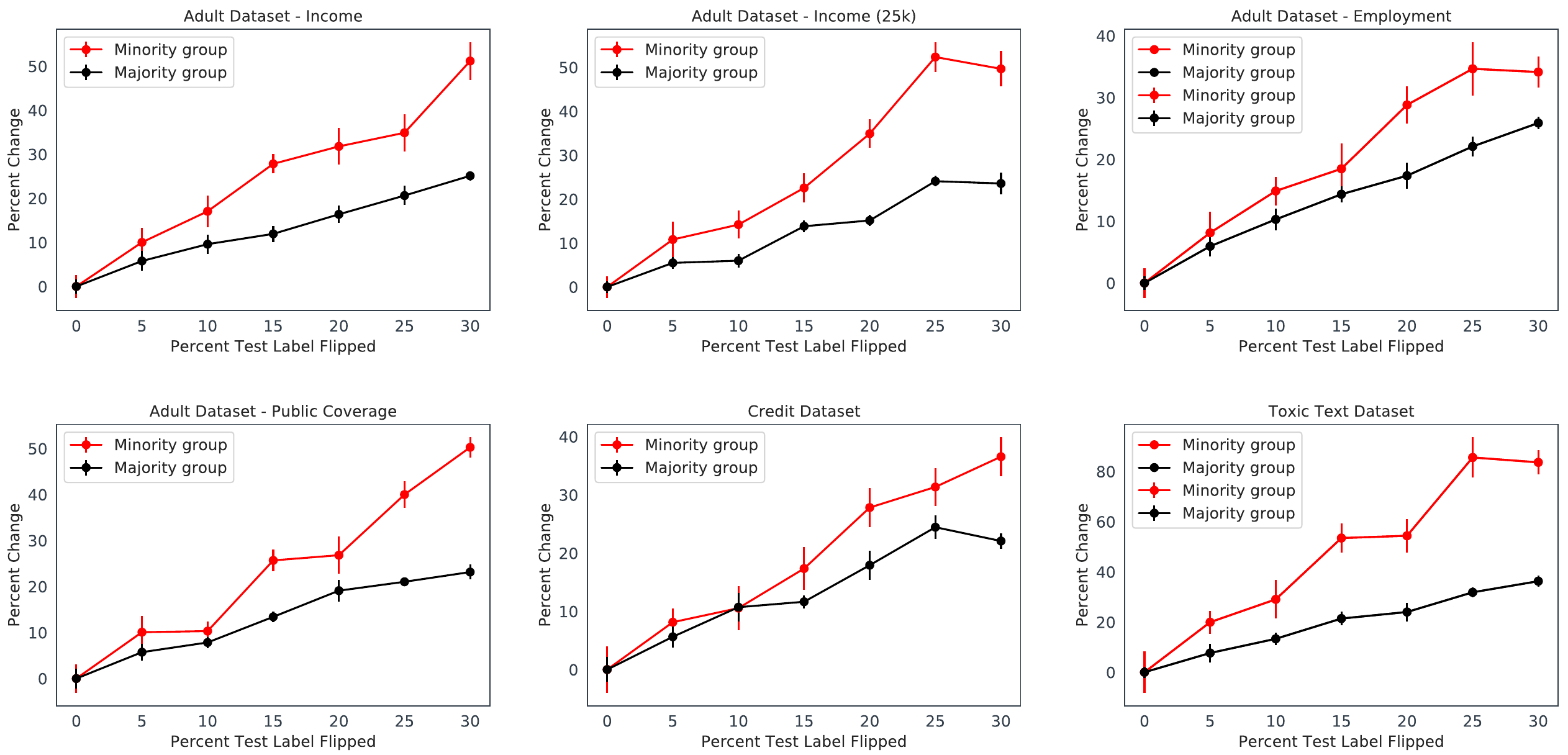}
\caption{\textbf{Test-time Label Flipping Results across}. For each dataset, we plot the percent change in calibration error versus the corresponding percentage change in label error. Here, we plot the minority (smallest) group as well as the majority (largest) group. These two groups represent two ends of the spectrum for the impact of label error. We observe that across all datasets, the minority group incurs higher percentage change in group calibration compared to the majority group.}
\label{fig:testlabelflipping}
\end{figure*}

\textbf{Results.} In Figure~\ref{fig:testlabelflipping}, we report results of the label flipping experiments across 6 tasks. On the horizontal axis, we have the percentage of labels flipped in the test dataset, while on the vertical axis, we have the percentage change in the model’s calibration. For each dataset, we compute model calibration for two demographic groups in the dataset, the majority and the minority—in size–groups. We do this since these two groups constitute the two ends of the spectrum in the dataset. As shown, we observe a more distinctive effect for the minority group across all datasets. This is to be expected since flipping even a small number samples in the minority group can have a dramatic effect on test and training accuracy within this group. For both groups, we observe a changes to the calibration error. For example, for the Income prediction task on the Adult dataset, a 10 percent label error induces at least a 20 percent change in the model’s test calibration error. These results suggest that test-time label error has more pronounced effects for minority groups. Similarly, we observe for other disparity metrics (See Appendix) across all model classes that increases in percentage of labels flipped disproportionately affects the minority group.

\subsection{Sensitivity to Training Label Error}
\textbf{Overview \& Experimental Setup.} The goal of the training-time empirical tests is to measure the impact of label error on a trained model. More specifically, given a training set in which a fraction of the samples’ labels have been flipped, what effect does the label error have on the calibration error compared to a model trained on data without label error? We simulate this setting by creating multiple copies of each of the datasets where a percentage of the training labels have been flipped uniformly at random. We then assess the model calibration of these different model using the same fixed test dataset. Under similar experimental training conditions for these models, we are then able to quantify the effect of training label error on a model’s test calibration error. We conduct this analysis across all dataset-model task pairs.

\begin{figure*}[!h]
\centering
\includegraphics[scale=0.25, page=4]{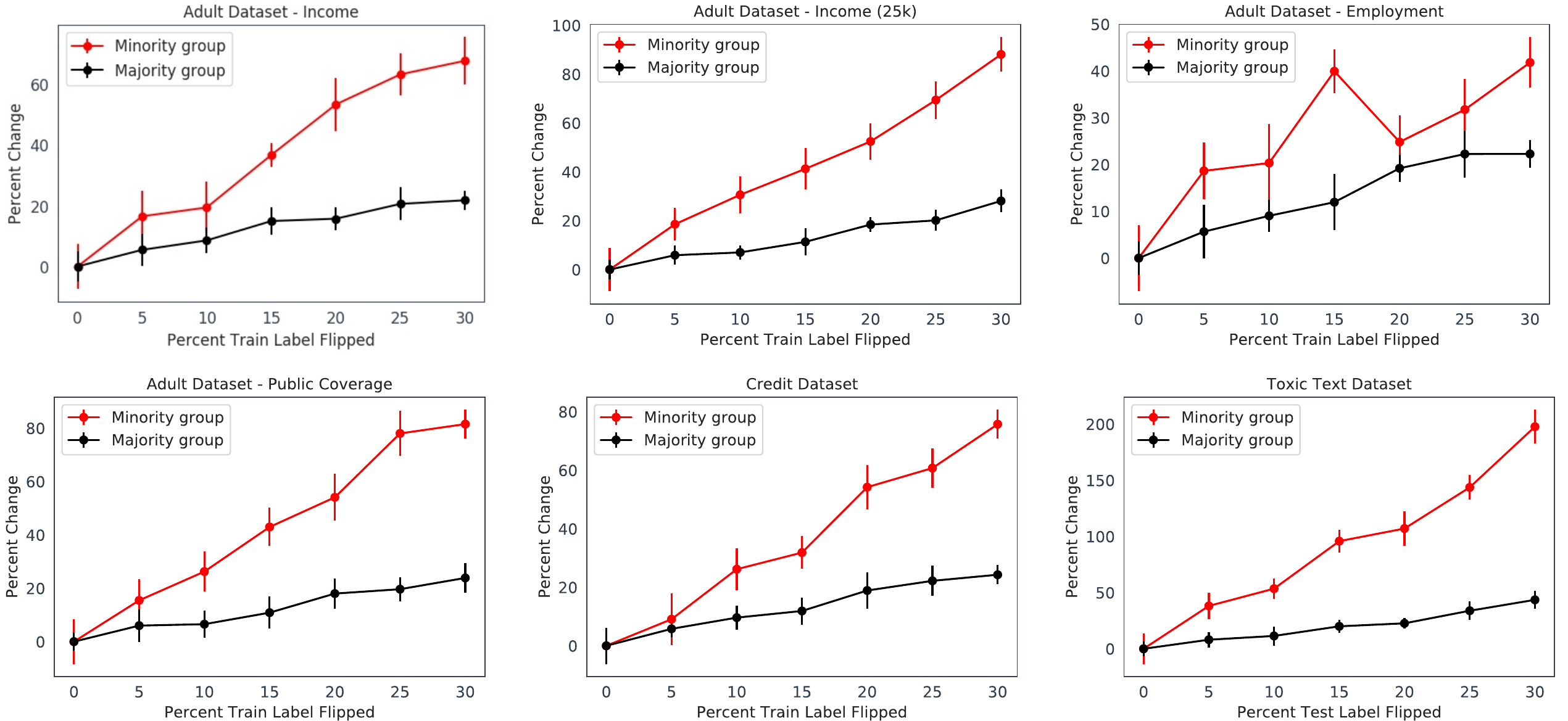}
\caption{\textbf{\textbf{Training-time Label Flipping Results}}. For each dataset, we plot the percent change in calibration error versus the corresponding percentage change in label error for the training set. Here, we plot the minority (smallest) group as well as the majority (largest) groups by size. Similar to the test-time setting, we observe that across all datasets, the minority group incurs higher percentage change in group calibration compared to the majority group. However, we observe a larger magnitude change for the minority groups.}
\label{fig:traininglabelflipping}
\end{figure*}

\textbf{Results \& Implications.} We show the results of the training-time experiments in Figure~\ref{fig:traininglabelflipping}. Similar to the test-time experiments, we find minority groups are more sensitive to label error than larger groups. Specifically, we find that even a 5 percent label error can induce significant changes in the disparity metrics, of a model trained on such data, for these groups. 

 A conjecture for the higher sensitivity to extreme training-time error is that a model trained on significant label error might have a more difficult time learning patterns in the minority class where there are not enough samples to begin with. Consequently, the generalization performance of this model worsens for inputs that belong to the minority group. Alternatively, in the majority group, the proportion of corrupted labels due to label error is smaller. This might mean that uniform flipping does not affect the proportion of true labels compared to the minority group. Even though the majority group exhibits label error, there still exists enough samples with true labels such that a model can learn the underlying signal for the majority class. 
 
 A second important finding is that overparameterization seems to confer more resilience to training label error. We find that for the same levels of training label error, an overparametrized model is less sensitive to such change compared to a model with a smaller number of parameters. Recent work suggests that models that learn functions that are more aligned with the underlying target function of the data generation process are more resilient to training label error~\citep{li2021does}. It might be that compared to linear and tree-based models, an overparametrized deep net is more capable of learning an aligned function.

 \subsection{Noise-Aware Robust Learning has Disparate Impact}
 \label{subsec:noiseaware}

 \begin{wrapfigure}{r}{0.3\textwidth}
\includegraphics[scale=0.25]{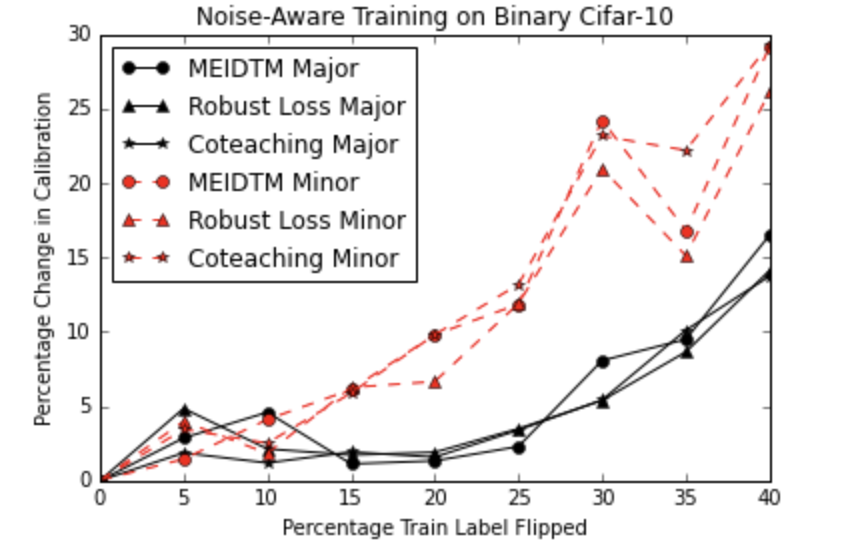}
\caption{\textbf{Effect of Noise-aware algorithms on group calibration}.}
\label{fig:noiseawaresensitivity}
\end{wrapfigure}

 \textbf{Overview \& Experimental Setup.} We now assess whether training models with noise-aware algorithmic interventions (e.g. robust loss functions~\citep{ma2020normalized, ghosh2017robust}) results in models whose disparity metrics have reduced sensitivity to label error in the training set. We test this hypothesis on a modified Cifar-10 dataset following the setting of ~\citet{hall2022systematic}. Specifically, the Cifar-10 dataset is modified to a binary classification setting along with group labels by inverting a subset of each class's examples. Given a specified parameter $\epsilon \in [0, 1/2]$, a $\frac{1}{2} - \epsilon$ of the negative class is inverted, while a $\frac{1}{2} + \epsilon$ of the positive class is inverted leading to $2\epsilon$ fraction of one group of samples and $1-2\epsilon$ of the other group. In all experiments we set $\epsilon=0.15$ for a 30 percent minority group membership. We replicate the label flipping experiment on the task with a Resnet-18 model. We test the MEIDTM~\citep{cheng2022instance}, DivideMix~\citep{li2020dividemix}, and a robust loss approach~\citep{ghosh2017robust}.

 \textbf{Results.}  At a high level, for the majority group, we find that group calibration remains resilient to low rates of label error (below 25 percent). At high rates (>30 percent label error), we start to see increased sensitivity. However, for the minority group (30 percent of the dataset), we observe group calibration remains sensitive to label error even at low levels. This finding suggests that noise-aware methods show are more effective for larger groups in the data. A similar observation has also been made for other algorithmic interventions like Pruning~\citep{tran2022pruning, hooker2019compressed}, differential privacy~\citep{bagdasaryan2019differential}, selective classification~\citep{jones2020selective} and adversarial training~\citep{xu2021robust}.

\section{Influence of Training Label on Test Disparity Metric}
\label{influence_method}
We now present an approach for estimating the `influence' of perturbing a training point's label on a disparity metric of interest. We consider: 1) up-weighting a training point, and 2) perturbing the training label.

\textbf{Upweighting a training point.}~Let $\hat{\theta}_{-z_i}$ be the ERM solution when a model is trained on all data points, $\{z_j\}_{j=1}^{n}$, except $z_i$. The influence, $\mathcal{I}_{\mathrm{up, params}}$, of datapoint, $z_i$, on the model parameters is then defined: $\hat{\theta}_{-z_i} - \hat{\theta}$. This measure indicates how much the parameters change when the model is `refit' on all training data points except $z_i$.~\citet{koh2017understanding} give a closed-form estimate of this quantity as:

\begin{align}
    \mathcal{I}_{\mathrm{up, params}} \defeq \frac{d\hat{\theta}_{\epsilon,~z_i}}{d\epsilon}\bigg\rvert_{\epsilon=0} = - H^{-1}_{\hat{\theta}}\nabla_{\theta}\ell(z_i, \hat{\theta}),
    \label{eqn:influenceparams}
\end{align}
where $H$ is the hessian, i.e., $H_{\hat{\theta}} \defeq \frac{1}{n}\sum^{n}_{i=1}\nabla^{2}_{\theta}\ell(z_i, \theta)$. 

The loss on a test example, $\ell(z_t, \hat{\theta})$, is a function of the model parameters, so using the chain-rule, we can estimate the influence, $\mathcal{I}_{\mathrm{up, loss}}(z_i, z_t)$, of a training point, $z_i$, on $\ell(z_t, \hat{\theta})$ as: 

\begin{align}
    \mathcal{I}_{\mathrm{up, loss}} (z_i, z_t) &\defeq \frac{d\ell(z_t, \hat{\theta}_{\epsilon,~z_i})}{d\epsilon}\bigg\rvert_{\epsilon=0} = - \nabla_\theta\ell(z_t, \hat{\theta})^\top H^{-1}_{\hat{\theta}}\nabla_{\theta}\ell(z_i, \hat{\theta}).
    \label{eqn:influenceloss}
\end{align}

\textbf{Perturbing a training point's label.} A second notion of influence that \citet{koh2017understanding} study is how perturbing a training point leads to changes in the model parameters. Specifically, given a training input, $z_i$, that is a tuple $(x_i, y_i)$, how would the perturbation, $z_i \rightarrow z_{i,\delta}$, which is defined as  $(x_i, y_i) \rightarrow (x_i, y_i + \delta)$, change the model's predictions?~\citet{koh2017understanding} give a closed-form estimate of this quantity as: 

\begin{align}
    \mathcal{I}_{\mathrm{pert, loss, y}}(z_j, z_t) 
    &\approx - \nabla_\theta\ell(z_t, \hat{\theta}_{z_{j, \delta}, -z_j})^\top H^{-1}_{\hat{\theta}}\nabla_y\nabla_\theta\ell(z_j, \hat{\theta}).
    \label{eqn:influencepertublosslabel}
\end{align}

\paragraph{Adapting influence functions to group disparity metrics.}
We now propose modifications that allow us to compute the influence of a training point on a test group disparity metric (See Appendix~\ref{influence_details_appendix} for longer discussion). Let $S_t$ be a set of test examples. We can then denote $\mathcal{GD}(S_t, \hat{\theta})$ as the group disparity metric of interest, e.g., the estimated ECE for the set $S_t$ given parameter setting $\hat{\theta}$. 

\textbf{Influence of upweighting a training point on a test group disparity metric.} A group disparity metric on the test set is a function of the model parameters; consequently, we can apply the chain rule to $\mathcal{I}_{\mathrm{up, params}}$ (from Equation~\ref{eqn:influenceparams}) to estimate the influence, $ \mathcal{I}_{\mathrm{up, disparity}}$, of up-weighting a training point on the disparity metric as follows: 
\begin{align}
    \mathcal{I}_{\mathrm{up, disparity}} (z_i, S_t) &\defeq \frac{d\mathcal{GD}(S_t, \hat{\theta}_{\epsilon,~z_i})}{d\epsilon}\bigg\rvert_{\epsilon=0}\nonumber = - \nabla_\theta\mathcal{GD}(S_t, \hat{\theta})^\top\frac{d\hat{\theta}_{\epsilon,~z_i}}{d\epsilon}\bigg\rvert_{\epsilon=0}, \nonumber \\
    &= - \nabla_\theta\mathcal{GD}(S_t, \hat{\theta})^\top H^{-1}_{\hat{\theta}}\nabla_{\theta}\ell(z_i, \hat{\theta}).
    \label{eqn:influencecalibration_text}
\end{align}
We now have a closed-form expression for a training point's influence on a test group disparity metric.

\textbf{Influence of perturbing a training point's label on a test group disparity metric.} We now consider the influence of a training label perturbation on a group disparity metric of interest. To do this, we simply consider the group disparity metric function as the quantity of interest instead of the test loss. Consequently, the closed-form expression for the influence of a modification to the training label on disparity for a given test set is:

\begin{align}
    \mathcal{I}_{\mathrm{pert, disparity, y}}(z_j, S_t) 
    &\approx - \nabla_\theta\mathcal{GD}(S_t, \hat{\theta})^\top H^{-1}_{\hat{\theta}}\nabla_y\nabla_\theta\ell(z_j, \hat{\theta}).
    \label{eqn:influencepertubcalibrationlabel_text}
\end{align}

With Equations~\ref{eqn:influencecalibration_text} and ~\ref{eqn:influencepertubcalibrationlabel_text}, we have the key quantities of interest that allows us to rank training points, in terms of influence, on the test group disparity metric. 

\section{Identifying and Correcting Training Label Error}
\label{influence_empirical}
In this section, we empirically assess the modified influence expressions for calibration across these datasets for prioritizing mislabelled samples. We find that the prioritization scheme shows improvement, compared to alternative approaches. In addition, we propose an approach to automatically correct the labels identified by our proposed approach.

\subsection{Identifying Label Error}
\label{identifylabelerror}

\textbf{Overview \& Experimental Question.} We are interested in surfacing training points whose change in label will induce a concomitant change in a test disparity metric like group calibration. Specifically, we ask: When the training points are ranked by influence on test calibration, are the most highly influential training points most likely to have the wrong labels? We conduct our experiments to directly measure a method's ability to answer this question.

\textbf{Experimental Setup.} For each dataset, we randomly flip the labels of $10-30$ percent of the training samples. We then train on this modified dataset.  In this task, we have direct access to the ground-truth of the exact samples whose labels were flipped. This allows us to directly compare the performance of our proposed methods to each of the baselines on this task. We then rank training points using a number of baseline approaches as well as the modified influence approaches. For the top $50$ examples, we consider what fraction of these examples had flipped labels in the training set. We discuss additional experimental details in the Appendix.

\textbf{Approaches \& Baselines.} We consider the following methods: 1) \textbf{IF-Calib}: The closed-form approximation to the influence of a training point on the test calibration; 2) \textbf{IF-Calib-Label}: The closed-form approximation to the influence of a training point's label on the test calibration; 3) \textbf{Loss}: A baseline method which is the training loss evaluated at each data point in the training set. The intuition is that, presumably, more difficult training samples will have higher training loss. We also consider several additional baselines that we discuss in the Appendix.


\begin{figure*}[!h]
\centering
\includegraphics[scale=0.3, page=5]{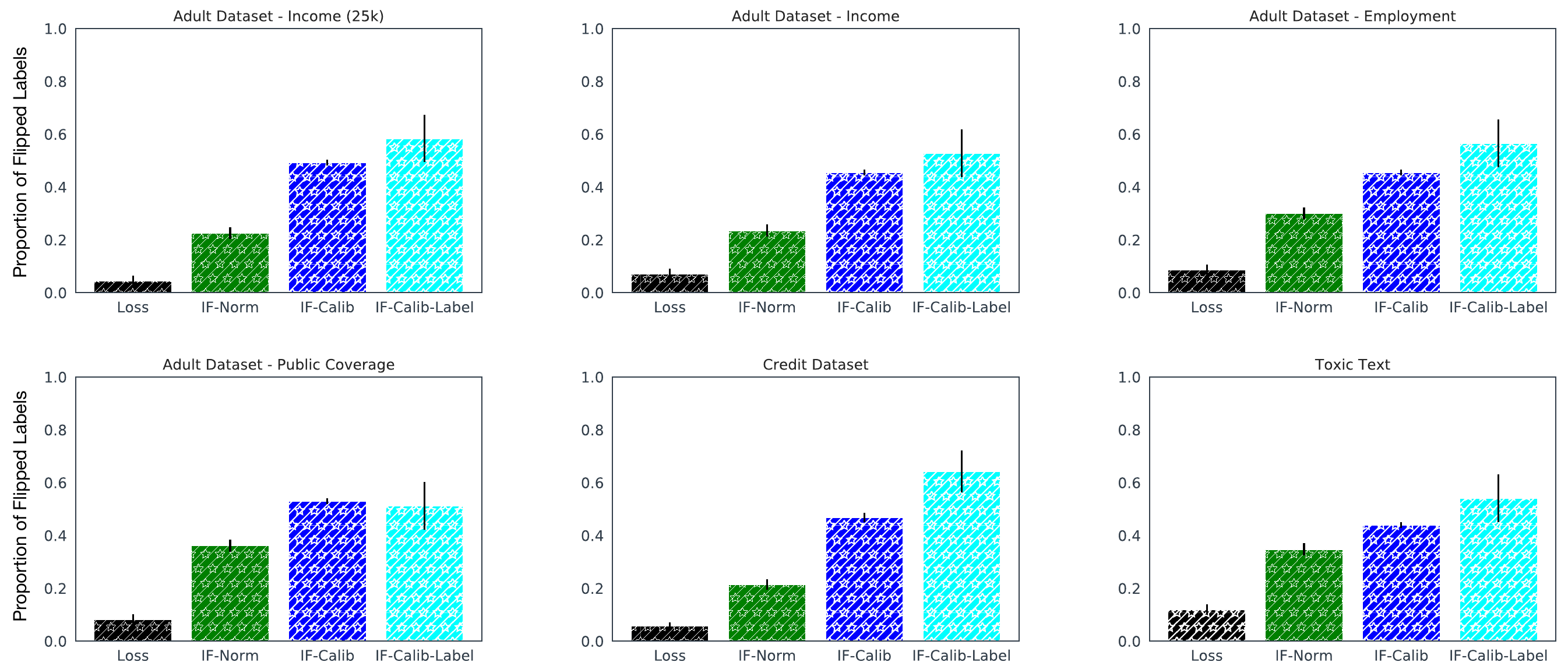}
\caption{\textbf{Empirical Results for Training Point Ranking Across 6 datasets}. 
For the top 50 most influential examples, we show the proportion of samples whose labels were flipped in the training data.}
\label{fig:labelinfluence}
\end{figure*}

\textbf{\textbf{Results: Prioritizing Samples.}} In Figure~\ref{fig:labelinfluence}, we show the performance of the two approximations that we consider in this work as well as two baselines. We plot the fraction of inputs, out of the top ranked $50$ ranked training points, whose labels were flipped in the training set. The higher this proportion, then the more effective an approach is in identifying the samples that likely have wrong labels. In practice, the goal is to surface these training samples and have a domain expert inspect them. If a larger proportion of the items to be inspected are mislabeled, then a higher proportion of training set mistakes, i.e. label error, can be fixed. Across the different datasets, we find a 10-40 percent improvement, compared to baseline approaches, in identifying critical training data points whose labels need to be reexamined.

We find the loss baseline to be ineffective for ranking in our experiments. A possible reason is that modern machine learning models can typically be trained to `memorize' the training data; resulting in settings where a model has low loss even on outliers or mislabeled examples. In such a case, ranking by training loss for a sample is an ineffective ranking strategy. We find that the noise-aware approaches perform similarly to the IF-Norm baseline. We defer the results of the uncertainty-based baselines and the noise-aware methods to Appendix (Section~\ref{appendix:rebuttalupdatesuncertaintyestimates}). We find that these baselines also underperform our proposed approaches.

\subsection{Correcting Label Error}
\label{correctlabelerror}
We take label error identification one step further to automatically relabelling inputs that have identified as critical. We restrict our focus to binary classification where the label set is $\{0, 1\}$, and the corresponding relabelling function is simply $1 - y_i$, where $y_i$ is the predicted label.

\textbf{Setup \& Experiment}: We consider the logistic regression model across all tasks for a setting with 20 percent training label error. We consider calibration as the disparity function of interest. We then rank the top 20 percent of training points by label-disparity influence, our proposed approach. For these points, we apply the relabelling function, and then fine-tune the model for an additional epoch with the modified labels. 

\textbf{Results:} First, we observe an improvement, in group calibration, across all groups, with larger improvement coming from the smallest group. As expected, we also observe a decrease in the average loss for the overall training set. These results point to increasing promise of automatic relabeling. 

\textbf{Theoretical Justification.} We now present a theorem that suggests that the influence priorization and relabeling scheme described above provably leads to better calibrated models.

\begin{theorem}
Given a $\kappa$-strongly convex loss function $\ell(., .)$, with $\kappa > 0$, a training dataset, $\mathcal{D}$, where $A$ indexes the data groups, and a model, $\hat{\theta} : x_i \rightarrow y_i$, optimized via $\ell(., .)$ that maps inputs to labels. Let $\mathcal{Q}$ be a set of test examples all belonging to group $A=a$, where $\mathrm{ECal}_{\mathcal{Q}}(\hat{\theta})$ is the expected calibration error of $\hat{\theta}$ on the set $\mathcal{Q}$. In addition, let $\mathcal{D}_{A=a}$ be the set of problematic training examples, belonging to group $a$, prioritized based on influence, i.e., $\mathcal{I}_{\mathrm{pert}, \mathrm{calib}, y^i}(x^i_a, \mathcal{Q}) > 0$. We term a model trained on a different training set ($\mathcal{D}_{+}$) where the problematic examples have been relabeled to be $\hat{\theta}_R$. Analogously, the expected calibration error of this new model on the set $\mathcal{Q}$ is $\mathrm{ECal}_{\mathcal{Q}}(\hat{\theta}_R)$. We have that:
\begin{center}
$\mathrm{ECal}_{\mathcal{Q}}(\hat{\theta}_R) \leq \mathrm{ECal}_{\mathcal{Q}}(\hat{\theta})$.
\end{center}
\label{thm:relabeledcalibration}
\end{theorem}

We defer the proof to the Appendix. Theorem~\ref{thm:relabeledcalibration} suggests that when a model is trained on a relabeled dataset, following the influence prioritization scheme, the expected group calibration of the retrained model should be lower than that of a model trained on a dataset that has not been relabeled.

\section{Related Work}
\label{related_work}
We discuss directly related work here, and defer a longer discussion to Section~\ref{appendix:relatedwork} of the Appendix.

\textbf{Impact of Label Error/Noise on Model Accuracy.} Learning under label error falls under the category more commonly known as \textit{learning under noise}~\citep{frenay2013classification, natarajan2013learning, bootkrajang2012label}. \textit{Noise} in learning can come from different either input features or the labels. In this work, we focus on label error---categorization mistakes associated with the label in both the test and training data. Previous work focused primarily on the effect of label error in the training data; however, we advance this line of work to investigate the effect of label error in the test data used to conduct a fairness audit on the reliability of the audit. Model resilience to training label error has been studied for both synthetic~\citep{arpit2017closer, zhang2021understanding, rolnick2017deep} and real-world noise settings~\citep{jiang2020beyond}. A major line of inquiry is the development of algorithmic approaches to learn accurate models given a training set with noisy labels. These approaches include model regularization~\citep{srivastava2014dropout, zhang2017mixup}, bootstrap~\citep{reed2014training}, knowledge distillation~\citep{jiang2020beyond}, instance weighting~\citep{ren2018learning, jiang2020identifying}, robust loss functions~\citep{ma2020normalized, ghosh2017robust}, or trusted data~\citep{hendrycks2018using}, joint training~\citep{wei2020combating}, mixture models in semi-supervised learning~\citep{li2020dividemix}, and methods to learn a transition matrix that captures noise dependencies~\citep{cheng2022instance}. In contrast to this line of work, we primarily seek to identify the problematic instances that need to be relabelled, often by a human labeler, and not automatically learn a model that is robust to label error. 

\textbf{Impact of Label Error on Model \textit{`Fairness'}.} This work contributes to the burgeoning area that studies the impact of label error on a model's `fairness' (termed `group-based disparity' in this paper) metrics.~\citet{fogliato2020fairness} studied a setting in which the labels used for model training are a noisy proxy for the true label of interest, e.g., predicting rearrest as a proxy for rearrest. \citet{wang2021fair} considers an ERM problem subject to group disparity constraints with group-dependent label noise, and provides theoretical results along with a scheme to obtain classifiers that are robust to noise. Different from their setting, we consider unconstrained ERM (no fairness constraints during learning). Similarly, \citet{konstantinov2021fairness} study the effect of adversarial data corruptions on fair learning in a PAC model. \citet{jiang2020identifying} propose a re-weighting scheme that is able to correct for label noise. 

\textbf{Influence Functions \& Their Uses.} Influence functions originate from robust statistics where it is used as a tool to identify outliers~\citep{cook1982residuals, cook1986assessment, hampel1974influence}.~\citet{koh2017understanding} introduced influence functions for modern machine learning models, and used them for various model debugging tasks. Most similar to our work,~\citet{sattigeri2022fair} and~\cite{li2022achieving} also consider the influence of a training point on model's disparity metric, and present intriguing results that demonstrate that reweighting training samples can improve a model's disparity metrics. Here, we focus specifically on the role of mislabeled examples; however, our goal aligns with theirs. Similarly, ~\citet{kong2021resolving} propose RDIA, a relabelling scheme based on the influence function that is able to provably correct for label error in the training data. RDIA identifies training samples that have a high influence on the test loss for a validation set; however, we focus on identifying training samples that influence a group-disparity metric on a test/audit set. We also rely on their technical results to prove Theorem~\ref{thm:relabeledcalibration}.

In recent work,~\citet{de2021leveraging} study expert consistency in data labeling and use influence functions to estimate the impact of labelers on a model's predictions. Along similar direction,~\citet{brunet2019understanding} adapt the influence function approach to measure how removing a small part of a training corpus, in a word embedding task, affects test bias as measured by the word embedding association test~\cite{caliskan2017semantics}.~\citet{feldman2020neural} use influence functions to estimate how likely a training point is to have been memorized by a model. More generally, influence functions are gaining widespread use as a tool for debugging model predictions~\citep{barshan2020relatif, han2020explaining, yeh2018representer, pruthi2020estimating}. Different from these uses of influence functions, here we isolate the effect of a training point's label on a model's disparity metric on a audit data.

\section{Conclusion}
In this paper, we sought to address two key questions: \textit{1) What is the impact of label error on a model's group disparity metric, especially for smaller groups in the data;} and \textit{2) How can a practitioner identify training samples whose labels would also lead to a significant change in the test disparity metric of interest?} We find that disparity metrics are, indeed, sensitive to test and training time label error particularly for minority groups in the data. In addition, we present an approach for estimating the `influence' of perturbing a training point's label on a disparity metric of interest, and find a 10-40\% improvement, compared to alternative approaches, in identifying training inputs that improve a model's disparity metric. We present an approach to estimate the effect of a training input's label on a model's group disparity metric. Lastly, perform a simple automatic relabel-and-finetune scheme that produces updated models with, provably, improved group calibration error.

Our findings come with certain limitations. In this work, we focused on the influence of label error on disparity metrics. However, other components of the ML pipeline can also impact downstream model performance. The proposed empirical tests simulate the impact of label error; however, it might be the case that real-world label error is less pernicious to model learning dynamics than the synthetic flipping results suggest. Ultimately, we see our work as helping to provide insight and as an additional tool for practitioners seeking to address the challenge of label error particularly in relation to a disparity metric of interest.

\bibliography{ref}
\bibliographystyle{plainnat}

\newpage

\appendix

\newpage 
\part{Appendix} 
\parttoc 

\section{Additional Related Work}
\label{appendix:relatedwork}

We discuss here additional literature. 

\textbf{Low Performance Subgroup Identification} We use influence functions to identify training points that have the most effect on the model's disparity metric; however, there are other approaches for surfacing training points that need to be prioritized. For example, \citet{kim2019multiaccuracy} propose an algorithm to identify groups in the data where a model has high test errors, and to `boost' the model performance for these groups. Similarly, \citet{creager2021environment} propose a two-stage scheme to identify critical subgroups in the data when demographic labels are not known ahead of time. Our approach differs from the aforementioned since we focus principally on the effect to the model's disparity metric instead of the test loss.

\textbf{Label Flipping.} We use labeling flipping as a primary tool in our empirical tests to measure the sensitivity of a model's disparity metrics to label error. Label flipping has also previously been used for alternative purposes~\cite{arpit2017closer, zhang2021understanding}. In now seminal work, \citet{zhang2021understanding} used label flipping and shuffling to show that deep neural networks easily memorize data with random labels. More generally, label flipping can also constitute an `adversarial attack' against an ML model, for which there are increasingly new methods to help defend against such attacks~\cite{rosenfeld2020certified}.

\paragraph{\textbf{Explainability}} Increasingly, `explanations' derived from a trained model can point to the reason why an input has been misclassified. Ultimately, one holy grail use-case for explanations has been help identify and suggest potential fixes for model performance disparities. Towards this end,~ \citet{pradhan2021interpretable} propose an approach that intervenes on the training data and measures the changes to downstream performance on the basis of these changes.

\section{Additional Discussion of Influence Functions}
\label{influence_appendix}
For completeness, we recap a non-rigorous derivation of the influence of a training point on the parameters of a model following ~\citet{koh2017understanding}. We first recall the notation and setup from the paper.

\textbf{Overview of Notation} We will consider prediction problems, i.e, settings where the task to to learn a mapping, $h$, from an input space $\mathcal{X} \in \mathbb{R}^d$ to an output space $\mathcal{Y} \in \mathbb{R}^k$. We follow the notation of \citet{koh2017understanding} in this work, so we consider the training points to be: $z_1, \ldots, z_n$, where each $z_i$ is a tuple $(x_i, y_i) \in \mathcal{X} \times \mathcal{Y}$. Given a function family, $\Theta$, the learning task to learn a particular parameter setting $\theta \in \Theta$. Throughout this work, we will only consider learning via empirical risk minimization (ERM), which corresponds to: $\hat{\theta} := \argmin_{\theta \in \Theta} \frac{1}{n}\sum_{i}^{n}\ell(z_i, \theta)$. Similar to \citet{koh2017understanding}, we will assume that the ERM solution is twice-differentiable and strictly convex in the parameters.

\textbf{Upweighting a training point} Let $\hat{\theta}_{-z_i}$ be the ERM solution when a model is trained on all data points except $z_i$. The influence, $\mathcal{I}_{\mathrm{up, params}}$, of datapoint, $z_i$, is then defined as the change: $\hat{\theta}_{-z_i} - \hat{\theta}$.

We can define the empirical risk as: 
\[
R(\theta) \defeq \frac{1}{n}\sum_{i=1}^n\ell(z_i, \theta).
\]

Since we assumed that the ERM solution is twice-differentiable and strictly convex in the parameters, we can specify the hessian as: 

\[
H_{\hat{\theta}} \defeq \frac{1}{n}\sum^{n}_{i=1}\nabla^{2}_{\theta}\ell(z_i, \theta) = \nabla^2 R(\theta)
\]

To start, let:

\[
\hat{\theta}_{\epsilon,~z_i} = \argmin_{\theta \in \Theta} {R(\theta) + \epsilon\ell(z_i, \theta)}.
\]

Then we can define the parameter change to be: $\Delta_\epsilon = \hat{\theta}_{\epsilon,~z_i}  - \hat{\theta}$. Since $\hat{\theta}$ does not depend on $\epsilon$, we have that:
\[ 
\frac{d\hat{\theta}_{\epsilon,~z_i}}{d\epsilon} = \frac{d\Delta_\epsilon}{d\epsilon}.
\]

We know that $\hat{\theta}_{\epsilon,~z_i}$ is a solution, i.e., minimizer, so we will form the first-order optimality conditions and form a Taylor expansion of that expression.

To do this we have that:

\begin{align*}
\nabla_\theta (R(\theta) + \epsilon\ell(z_i, \theta)) = 0, \\
\nabla R(\theta) + \epsilon\nabla\ell(z_i, \theta) = 0.
\end{align*}

The Taylor expansion of the above expression is then: 
\begin{align*}
[\nabla(R(\hat{\theta}) + \epsilon\nabla\ell(z_i, \theta)] + [\nabla^2 (R(\hat{\theta}) + \epsilon\nabla^2\ell(z_i, \theta)\Delta_\epsilon] \approx 0.
\end{align*}

Above, we only keep the first two terms of the Taylor Expansion.

We will now solve for $\Delta_\epsilon$ in the above equation and then differentiate by $\epsilon$ to obtain the final expression.

Solving for $\Delta_\epsilon$, results in: 
\begin{align*}
\Delta_\epsilon \approx -[\nabla^2 (R(\hat{\theta}) + \epsilon\nabla^2\ell(z_i, \theta)]^{-1} [\nabla(R(\hat{\theta}) + \epsilon\nabla\ell(z_i, \theta)].
\end{align*}

If we look at the expression for $\Delta_\epsilon$, we find that $\nabla(R(\hat{\theta})$ is zero, since we know that $\hat{\theta}$ minimizes the empirical risk $R$. In looking at the term $[\nabla^2 (R(\hat{\theta}) + \epsilon\nabla^2\ell(z_i, \theta)]$, we find that it is equivalent to $[H_{\hat{\theta}} + \epsilon\nabla^2\ell(z_i, \theta)]$. We can further make the assumption that the contribution of the $\epsilon\nabla^2\ell(z_i, \theta)$ term is small, so we have $[H_{\hat{\theta}} + \epsilon\nabla^2\ell(z_i, \theta)] \approx H_{\hat{\theta}}$. 

With the above assumptions and substitutions, we arrive at the new expression for $\Delta$, which is now: 
\begin{align*}
\Delta_\epsilon \approx -H_{\hat{\theta}}^{-1}\nabla\ell(z_i, \theta)\epsilon.
\end{align*}

Let's differentiate the above expression by $\epsilon$, and we have that: 

\[ 
\frac{d\Delta_\epsilon}{d\epsilon} = -H_{\hat{\theta}}^{-1}\nabla\ell(z_i, \theta).
\]

Now recall that: 
\[ 
\frac{d\hat{\theta}_{\epsilon,~z_i}}{d\epsilon} = \frac{d\Delta_\epsilon}{d\epsilon}, 
\]

which means: 

\[ 
\frac{d\hat{\theta}_{\epsilon,~z_i}}{d\epsilon} = \frac{d\Delta_\epsilon}{d\epsilon} = -H_{\hat{\theta}}^{-1}\nabla\ell(z_i, \theta).
\]

As discussed in the main draft, $\mathcal{I}_{\mathrm{up, params}}$ is defined to be the influence of estimate for training point $z_i$, so :

\[
\mathcal{I}_{\mathrm{up, params}} \defeq \frac{d\hat{\theta}_{\epsilon,~z_i}}{d\epsilon} = -H_{\hat{\theta}}^{-1}\nabla\ell(z_i, \theta).
\]

As we have seen, we obtain the influence of a training point on a the loss of a test simply by applying the chain rule to the loss-derivative quantity of interest. We will now extend this notion of influence.

\paragraph{\textbf{Upweighting a training point}} Let $\hat{\theta}_{-z_i}$ be the ERM solution when a model is trained on all data points except $z_i$. The influence, $\mathcal{I}_{\mathrm{up, params}}$, of datapoint, $z_i$, is then defined as the change: $\hat{\theta}_{-z_i} - \hat{\theta}$. This measure indicates how much the parameters change when the model is `refit' on all training data points except $z_i$. One approach to estimate this quantity is to train two copies of the model: one with all data points, and another on all data points except $z_i$ and then compare these two parameter settings. Such approach is time and compute prohibitive especially if one is interested in computing the influence of all training points---computing the influence of $n$ training points will require refiting the model $n$ times. For model classes that require significant compute to estimate, such a procedure is infeasible. To circumvent the retraining challenge, \citet{koh2017understanding} give a closed-form approximation to the influence quantity as:

\begin{align}
    \mathcal{I}_{\mathrm{up, params}} \defeq \frac{d\hat{\theta}_{\epsilon,~z_i}}{d\epsilon}\bigg\rvert_{\epsilon=0} = - H^{-1}_{\hat{\theta}}\nabla_{\theta}\ell(z_i, \hat{\theta}),
    \label{eqn:influenceparams_appendix}
\end{align}
where $H$ is the hessian of the loss and defined as: $H_{\hat{\theta}} \defeq \frac{1}{n}\sum_{n}^{i=1}\nabla^{2}_{\theta}\ell(z_i, \theta)$. Equation~\ref{eqn:influenceparams} gives a closed-form approximation of the influence of a training point $z_i$ on the model parameters. Recall that this quantity tells us how the model parameters change when the training point is upweighted by a small amount (say $\epsilon$). Upweighting a training point by $-\frac{1}{n}$ is equivalent to removing this training point from the dataset.

Given a closed-form approximation to $\mathcal{I}_{\mathrm{up, params}}$, we can estimate the influence of a training point on functions of the parameters. For example,~\citet{koh2017understanding} show, using a chain-rule argument, that the influence, $\mathcal{I}_{\mathrm{up, loss}}(z_i, z_t)$, of a training point, $z_i$, on the test loss for a test example has the following closed-form expression: 

\begin{align}
    \mathcal{I}_{\mathrm{up, loss}} (z_i, z_t) &\defeq \frac{d\ell(z_t, \hat{\theta}_{\epsilon,~z_i})}{d\epsilon}\bigg\rvert_{\epsilon=0} \nonumber \\
    &= - \nabla_\theta\ell(z_t, \hat{\theta})^\top\frac{d \hat{\theta}_{\epsilon,~z_i}}{d\epsilon}\bigg\rvert_{\epsilon=0}, \nonumber \\
   &= - \nabla_\theta\ell(z_t, \hat{\theta})^\top H^{-1}_{\hat{\theta}}\nabla_{\theta}\ell(z_i, \hat{\theta}).
    \label{eqn:influenceloss_appendix}
\end{align}

As we have seen, we obtain the influence of a training point on a the loss of a test simply by applying the chain rule to the loss-derivative quantity of interest. We will now extend this notion of influence.

\paragraph{\textbf{Perturbing a training point}} A second notion of influence that \citet{koh2017understanding} study is how perturbing a training point leads to changes in the model parameters. Specifically, given a training input, $z_i$, that is a tuple $(x_i, y_i)$, then how would the perturbation, $z_i \rightarrow z_{i,\delta}$, which is defined as  $(x_i, y_i) \rightarrow (x_i + \delta, y_i)$, change the model's predictions? The key issue here is how an infinitesimal change in the input example changes the model parameters and predictions. Let $\hat{\theta}_{\epsilon, z_{i,\delta}, -z_j}$ be the ERM solution to the following minimization problem:
\[
\argmin_{\theta \in \Theta} \frac{1}{n}\sum_{i=1}^n \ell(z_i, \theta) + \epsilon\ell(z_{j,\delta}, \theta) - \epsilon\ell(z_{j}, \theta).
\]
As shown, $\hat{\theta}_{\epsilon, z_{i,\delta}, -z_j}$ is the ERM solution obtained when an infinitesimal mass, $\epsilon$, is shifted from $z_j$ i.e.  $(x_j, y_j)$ to $z_{j,\delta}$, i.e. $(x_j + \delta, y_j)$. Similarly, \citet{koh2017understanding} show that the closed-form approximation for such training point perturbation on the parameters is:

\begin{align}
    \mathcal{I}_{\mathrm{up, params}}(z_{j,\delta}) -  \mathcal{I}_{\mathrm{up, params}}(z_{j}) &= \frac{d\hat{\theta}_{\epsilon, z_{i,\delta}, -z_j}}{d\epsilon}\bigg\rvert_{\epsilon=0},\nonumber\\
    &\approx -H^{-1}_{\hat{\theta}}[\nabla_x\nabla_\theta\ell(z_j, \hat{\theta})]\delta.
    \label{eqn:influencepertub_appendix}
\end{align}

With a closed-form expression for the influence of perturbing a training point, we can also obtain similar forms for functions of the parameters. Consequently, to obtain the influence of perturbing a training point on the model's prediction (i.e. a model with parameters $\hat{\theta}$), we differentiate with respect to $\delta$, and apply the chain rule to obtain:

\begin{align}
    \mathcal{I}_{\mathrm{pert, loss}}(z_j, z_t) &\defeq \nabla_\delta\ell(z_t, \hat{\theta}_{z_{j, \delta}, -z_j} )\bigg\rvert_{\delta=0} \nonumber \\
    &\approx - \nabla_\theta\ell(z_t, \hat{\theta}_{z_{j, \delta}, -z_j})^\top H^{-1}_{\hat{\theta}}\nabla_x\nabla_\theta\ell(z_j, \hat{\theta}).
    \label{eqn:influencepertubloss_appendix}
\end{align}

We now have closed-form expressions for estimating the influence of a training point and a modification to this training point on the loss of a new test example respectively. Shortly, we will make simple modifications to these equations to obtain the influence of the training point on the calibration for a test point.

\paragraph{\textbf{Perturbing a training point's label}} We are also interested in how a modification to the \textit{training label} influences the parameters and the test loss. In this case, we are interested in how the perturbation, $(x_i, y_i) \rightarrow (x_i, y_i + \delta)$ changes the model's predictions. An analogous calculation to that of Equation~\ref{eqn:influencepertub_appendix} results in:

\begin{align}
    \mathcal{I}_{\mathrm{up, params, y}}(z_{j,\delta}) -  \mathcal{I}_{\mathrm{up, params, y}}(z_{j}) &\approx -H^{-1}_{\hat{\theta}}[\nabla_y\nabla_\theta\ell(z_j, \hat{\theta})]\delta.
    \label{eqn:influencepertublabel_appendix}
\end{align}

In Equation~\ref{eqn:influencepertublabel_appendix}, the inner gradient is taken with respect to the label, $y$, as opposed to the the input, $x$, as was done in Equation~\ref{eqn:influencepertub_appendix}. Consequently, the closed-form expression for the influence of a modification to the training label on the loss of a new test example is:

\begin{align}
    \mathcal{I}_{\mathrm{pert, loss, y}}(z_j, z_t) 
    &\approx - \nabla_\theta\ell(z_t, \hat{\theta}_{z_{j, \delta}, -z_j})^\top H^{-1}_{\hat{\theta}}\nabla_y\nabla_\theta\ell(z_j, \hat{\theta}).
    \label{eqn:influencepertublosslabel_appendix}
\end{align}

We now have all the closed-form expressions we need for estimating influence with respect to a disparity metric of interest.

\paragraph{Adapting influence functions to group disparity metrics.}
We now propose modifications that allow us to compute the influence of a training point on a test group disparity metric. Let $S_t$ be a set of test examples. We can then denote $\mathcal{GD}(S_t, \hat{\theta})$ as the group disparity metric of interest, e.g., the estimated ECE for the set $S_t$ given parameter setting $\hat{\theta}$. 

\textbf{Influence of upweighting a training point on a test group disparity metric.} A group disparity metric on the test set is a function of the model parameters; consequently, we can apply the chain rule to $\mathcal{I}_{\mathrm{up, params}}$ (from Equation~\ref{eqn:influenceparams}) to estimate the influence, $ \mathcal{I}_{\mathrm{up, disparity}}$, of up-weighting a training point on the disparity metric as follows: 
\begin{align}
    \mathcal{I}_{\mathrm{up, disparity}} (z_i, S_t) &\defeq \frac{d\mathcal{GD}(S_t, \hat{\theta}_{\epsilon,~z_i})}{d\epsilon}\bigg\rvert_{\epsilon=0}\nonumber = - \nabla_\theta\mathcal{GD}(S_t, \hat{\theta})^\top\frac{d\hat{\theta}_{\epsilon,~z_i}}{d\epsilon}\bigg\rvert_{\epsilon=0}, \nonumber \\
    &= - \nabla_\theta\mathcal{GD}(S_t, \hat{\theta})^\top H^{-1}_{\hat{\theta}}\nabla_{\theta}\ell(z_i, \hat{\theta}).
    \label{eqn:influencecalibration}
\end{align}
We now have a closed-form expression for a training point's influence on a test group disparity metric.

\textbf{Influence of perturbing a training point's label on a test group disparity metric.} We now consider the influence of a training label perturbation on a group disparity metric of interest. To do this, we simply consider the group disparity metric function as the quantity of interest instead of the test loss. Consequently, the closed-form expression for the influence of a modification to the training label on disparity for a given test set is:

\begin{align}
    \mathcal{I}_{\mathrm{pert, disparity, y}}(z_j, S_t) 
    &\approx - \nabla_\theta\mathcal{GD}(S_t, \hat{\theta})^\top H^{-1}_{\hat{\theta}}\nabla_y\nabla_\theta\ell(z_j, \hat{\theta}).
    \label{eqn:influencepertubcalibrationlabel}
\end{align}

With Equations~\ref{eqn:influencecalibration} and~\ref{eqn:influencepertubcalibrationlabel}, we have the key quantities of interest that allows us to rank training points, in terms of influence, on the test group disparity metric. We can then use these quantities to prioritize samples that should be more carefully inspected.


\section{Appendix: Datasets, Models, \& Experimental Details}
We provide additional details about the datasets, models, and general experimental details. We follow the discussion in the background section of the paper and provide additional details where necessary. We plan to release opensource code that can be used to replicate all of or analyses.

\subsection{Datasets}
\label{dataset_appendix}
We start with a discussion of the datasets used in this work.
\paragraph{Text Toxicity Classification} we use a publicly available dataset from JIGSAW called the \textit{unintended bias in toxicity classification} dataset. The TC dataset contains a subset of comments from the `Civil Comments' platform that have been annotated by human raters for level of toxic severity. For example, a comment can be tagged as `benign', `obscene', `threat', `insult', or `sexually explicit' among several categories. A given text is then indicated as `toxic' or `not toxic' on the basis of these tags. The task here is a binary classification one, which is to categorize an input text as either toxic or not. 

To obtain the toxicity labels, each comment was shown to up to 10 annotators. Annotators were asked to:``Rate the toxicity of this comment": Very Toxic, Toxic, Hard to Say, and Not Toxic. These ratings were then aggregated with the target value representing the fraction of annotations that annotations fell within the former two categories.

To collect the identity labels, annotators were asked to indicate all identities that were mentioned in the comment. An example question that was asked as part of this annotation effort was:``What genders are mentioned in the comment?": Male, Female, Transgender, Other Gender, and No gender mentioned. We consider the Gender variable to be the sensitive attribute of interest. 

For the dataset, we taken of the raw sentence text and convert them into embedding vectors using the XLM-R models obtaining a vector, for each sentence, that is $768$ dimensional. We then further reduce the input dimension of this vector from $768$ to $50$ via a random projection. We make this reduction to make computing the hessian of the loss function of the logistic regression model trained on this data easy to compute. The $50$-dimensional embedding is then used as input for our analyses.

\paragraph{Adult Census Dataset} we use a series of tabular datasets more broadly called the `Adult Dataset'. Specifically, we consider a recently revamped version of the dataset introduced by~\citet{ding2021retiring}, which is derived from the broader US Census. We consider the following tasks among the compilation of datasets available in this database:
    \begin{enumerate}
        \item ACSIncome Prediction, where the task is to predict whether an individual has an income above 50000 USD; 
        \item ACSIncome Prediction (25k) where the task is to predict whether an individual has an income above 25000 USD;
        \item ACSEmployment Prediction, where the task is to predict whether the adult individual is employed; and
        \item ACSPublic Coverage: where the task is to predict whether a low-income individual has coverage from public health insurance.
    \end{enumerate}
 The adult dataset comes annotated with race categories, which we take as the key demographic variable in our analyses. We restrict our analyses to data from year 2018 for the state of California across all datasets.

\paragraph{Credit Dataset}: a dataset of financial transactions that consists of aggregate demographic and credit information for customers of a large commercial bank. For each customer, information includes their gender, marital status, educational level, employment status, income, age, and asset. Using this information, the bank estimates probability of default on loan for each customer (a scalar between 0 and 1). We consider the prediction of the default probability as our primary focus. The dataset comes with gender as a sensitive attribute for two categories: Male and Female. While this dataset is not publicly available---it is hosted by the first author's institution, it has been used as part of previous work studying financial well-being classification.

\subsection{Models}.
\label{models_appendix}
\paragraph{Models}  For the logistic regression model. All but one of the datasets we consider are tabular, and mostly low-dimensional. We implement the logistic regression model in PyTorch. We tuned the batch size using the validation set in the range [16, 32, 64, 128] across all datasets, but did not observe substantial across differences, so we set default batch size to be: 128. We use the SGD plain optimizer with a default learning rate of $0.001$, we train the all models for 20 epochs. For the Resnet-18 and GBT models, we keep the default parameter settings.

\subsection{Additional Discussion on ECE and Brier Score}
The principal metric that we consider in this work is group calibration. We also consider other group-based metrics such as true-positive and false negative rates. Let $\hat{y_i} = h(x_i)$ be the output of a trained model of interest for input $x_i$. The output can either be an output probability for a binary prediction or a class prediction in a multi-class setting. We denote the ground-truth confidence or probability of correctness as $\hat{p_i}$. We say $h$ is calibrated if $\hat{p_i}$ represents a true probability. As an example, if given 100 predictions with confidence $0.95$, then if $h$ is calibrated, $95$ of these predictions should be correct given ground-truth labels.

Several recent works have also studied model calibration and found that modern neural network models are uncalibrated. \citet{guo2017calibration} found that a simple temperature based Platt-scaling post processing was the most effective technique. While the literature abound with disparity metrics, several of which provide conflicting and often counter-intuitive insights simultaneously, group calibration has emerged as a useful metric in practice~\cite{pleiss2017fairness}. 

There are a variety of metrics for quantifying a model's level of calibration. In this work, we measure calibration with using the metric: Expected Calibration Error (ECE)~\cite{naeini2015obtaining}. A calibration metric summarizes the difference between the empirical distribution of a model's prediction and the ground-truth probabilities for a perfectly calibrated classifier. Here, perfect calibration is defined as:

\begin{align*}
\mathbb{P}\left( \hat{y}=y \vert \hat{p} = p \right), \forall p \in [0, 1].
\end{align*}

In practice, the quantity above is impossible to compute, so previous literature made empirical approximations. We follow binning the approach by \citet{guo2017calibration}. To estimate the accuracy empirically, we  group predictions into $M$ interval bins (each of size 1/M) and calculate the accuracy of each bin. Across all experiments, we take $M=10$---recent work~\cite{Kueppers_2020_CVPR_Workshops} found this default setting effective across a range of datasets. We discuss this choice in more detail in the appendix. Let $B_m$ be the set of indices of samples whose prediction confidence falls into the interval $I_m = [\frac{m-1}{M}, \frac{m}{M}]$. Here the accuracy of $B_m$ is then: 

\begin{align*}
\mathrm{acc(B_m)}= \frac{1}{\vert B_m \vert} \sum_{i \in B_m} \mathbbm{1}(\hat{y_i} = y_i),
\end{align*}
where $\hat{y_i}$, and $y_i$ are the predicted and true class labels for data sample $i$.  

The ECE is the absolute difference in expectation between confidence and accuracy. We can define the average confidence  within a bin $B_m$ as: 

\begin{align*}
\mathrm{conf(B_m)}= \frac{1}{\vert B_m \vert} \sum_{i \in B_m} \hat{p_i},
\end{align*}
where $\hat{p_i}$ is the confidence for sample $i$. Consequently, a perfectly calibrated model will have accuracy and confidence equal for all bins.

A notion of miscalibration is then which is the difference in expectation between confidence and accuracy.

The ECE approximates the above notion of miscalibration, and is defined as: 
\begin{align*}
 \sum_{m=1}^M \frac{\vert B_m \vert}{n} \vert \mathrm{conf(B_m)} - \mathrm{acc(B_m)} \vert.
\end{align*}
We use ECE as the primary metric of miscalibration in this work. As we will see, we will compare ECE metrics across various data groups to help identify input groups where a model is miscalibrated.

Throughout all of our experiments, we use a logistic regression model. All but one of the datasets we consider are tabular, and mostly low-dimensional, so we did not observe substantial accuracy gains for more sophisticated models. In addition, as we will see, our proposed influence function approach requires Hessian-vector products, which are more easily tractable for lower-dimensional models; in the case of the logistic regression model, the hessian of the loss has a simple closed-form expression that is easy to work with and compute. Lastly, a point we will return to in the discussion section is that influence function approaches have been shown to be challenging to work with for modern deep learning models.

For the TC dataset, we obtain embedding vectors, $768$ dimensional, for all samples using the XLM-R, a state of the art multi-lingual model~\cite{conneau2019unsupervised}. We then further reduce the input dimension of this vector from $768$ to $50$ via a random projection. We make this reduction to make computing the hessian of the loss function of the logistic regression model trained on this data easy to compute. The $50$-dimensional embedding is then used as input for our analyses. Across all datasets, we take 70 percent of each dataset for training and model validation, while we keep 30 percent as a holdout set for computing the disparity metrics. For the 70 percent portion, we reserve 20 percent as validation set. We refer to the Appendix for additional details.

\paragraph{Compute.} All of the experiments in the paper we performed on a cluster with a single Tesla T4 GPU.

\section{Influence function Implementation}
\label{influence_details_appendix}
We perform our influence function experiments only on the logistic regression model class. We make this choice for scalability reasons---computing inverse hessian-vector products for the logistic regression model is straightforward and does not incur approximation challenges.

We present additional results here: 

\begin{table*}[t]
\centering
\begin{tabular}{lrrl}
Approach         & Adult & Credit       & Toxic  \\
\hline
IF-Disparity        &  $0.47$    & $0.51$  & $0.45$  \\
IF-Disparity-label           &  $0.71$    & $0.69$   & $0.82$ \\
IF-Norm         &  $0.24$    & $0.3$   & $0.27$ \\
Loss           &  $0.11$   & $0.21$  & $0.15$ \\
\end{tabular}
\caption{Additional influence function metrics across datasets.}
\label{tab:data-characteristics}
\end{table*}

\section{Additional Empirical Results}
\label{appendix:additionalempiricalresults}
\begin{figure*}[!h]
\centering
\includegraphics[scale=0.75]{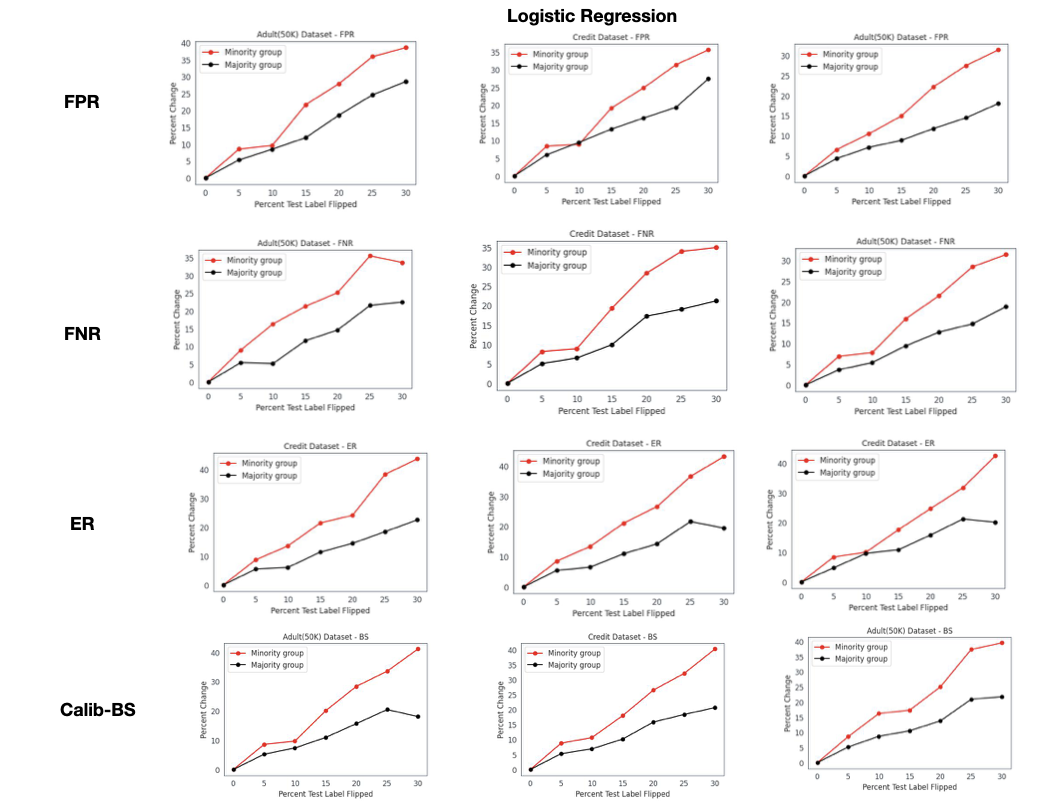}
\caption{\textbf{\textbf{Label Flipping Results for Logistic Regression Model Class}}. For each dataset, we plot the percent change in disparity metric versus the corresponding percentage change in label error for the test \& training set. Here, we plot the minority (smallest) group as well as the majority (largest) groups by size.}
\label{fig:appendixadditionaleresults1}
\end{figure*}

\begin{figure*}[!h]
\centering
\includegraphics[scale=0.75]{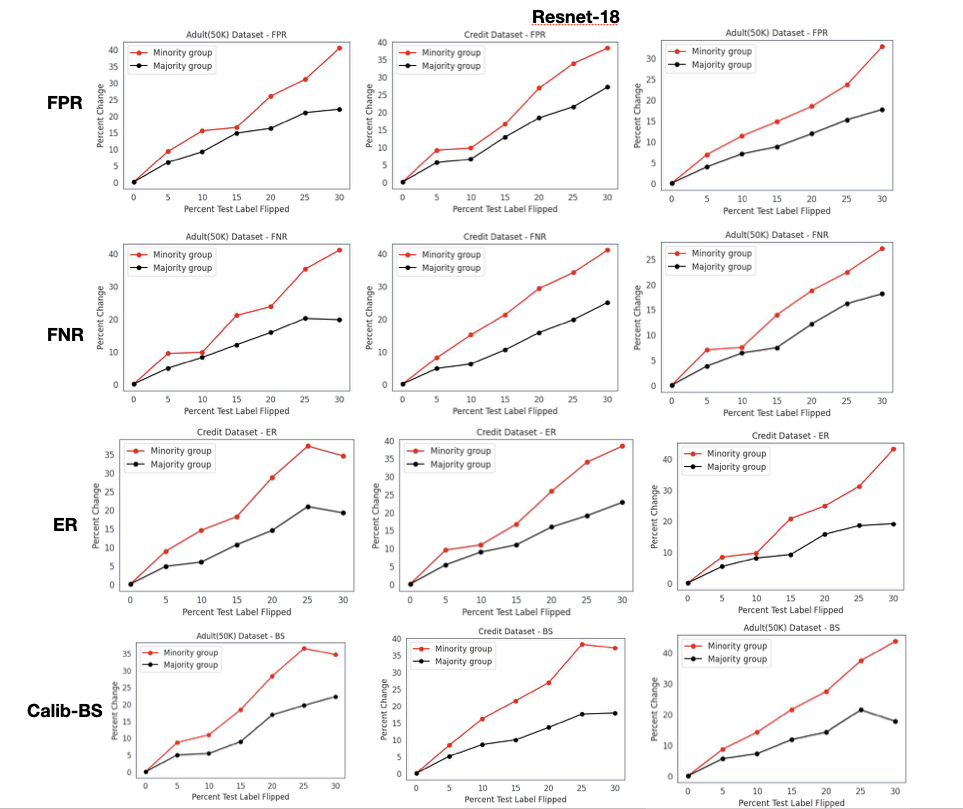}
\caption{\textbf{\textbf{Label Flipping Results for Resnet-18 Model Class}}. For each dataset, we plot the percent change in disparity metric versus the corresponding percentage change in label error for the test \& training set. Here, we plot the minority (smallest) group as well as the majority (largest) groups by size.}
\label{fig:appendixadditionaleresults1}
\end{figure*}

\begin{figure*}[!h]
\centering
\includegraphics[scale=0.75]{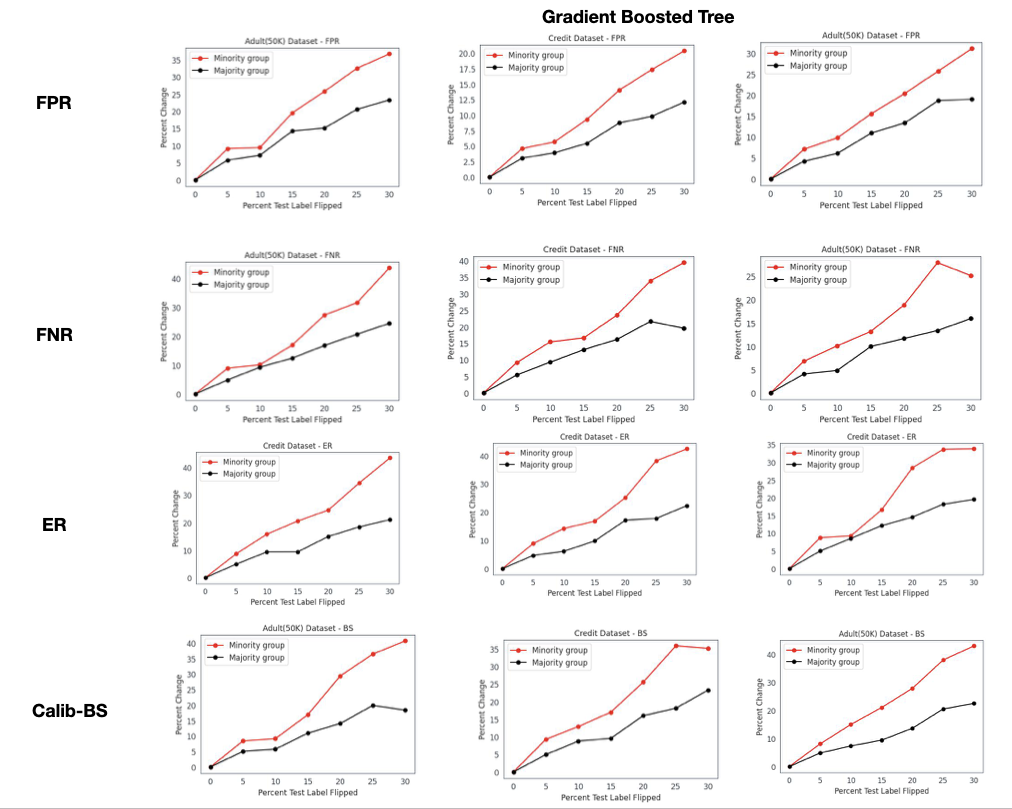}
\caption{\textbf{\textbf{Label Flipping Results for Gradient Boosted Regression Model Class}}. For each dataset, we plot the percent change in disparity metric versus the corresponding percentage change in label error for the test \& training set. Here, we plot the minority (smallest) group as well as the majority (largest) groups by size.}
\label{fig:appendixadditionaleresults1}
\end{figure*}

\begin{figure*}[!h]
\centering
\includegraphics[scale=0.75]{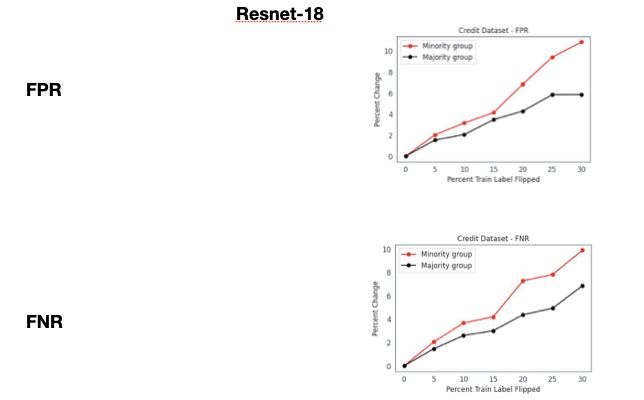}
\caption{\textbf{\textbf{Training Label Flipping Results for Resnet-18 Model Class}}.}
\label{fig:appendixadditionaleresults1}
\end{figure*}

\clearpage
\section{Noisy Example Identification: BenchMarking Uncertainty Estimation \& Noise-Aware Schemes}
\label{appendix:rebuttalupdatesuncertaintyestimates}
In this section, we benchmark two kinds of uncertainty estimation algorithms: 1) that estimates the uncertainty in a sample's label based on a  k-fold (5, and 1) cross validation score and 2) a logit based uncertainty estimation algorithm due to~\cite{wu2021logit}. We test these three variants under the sample setup as described in the main draft in Figure~4.

\begin{center}
\begin{tabular}{||c c c c c c||} 
 \hline
Method & Adult (Income 25k) & Adult (Income) & Adult (Employment) & Adult PC & Credit Dataset\\ [0.5ex] 
 \hline\hline
 Logit & 0.15 (0.05) & 0.17 (0.09) & 0.24 (0.09) & 0.27 (0.01) & 0.21(0.04)\\ 
 \hline
 CV ($k=1)$ & 0.21 (0.01) & 0.23 (0.05) & 0.25 (0.07) & 0.31 (0.11) & 0.19 (0.1) \\
 \hline
CV ($k=5)$ & 0.09 (0.06) & 0.11 (0.09) & 0.17 (0.1) & 0.23 (0.14) & 0.085 (0.02) \\
 \hline
Confident Learning & 0.17 (0.1) & 0.18 (0.12) & 0.2 (0.039) & 0.25 (0.11) & 0.15 (0.05) \\
 \hline
 MEIDTM & 0.165 (0.03) & 0.17 (0.025) & 0.19 (0.11) & 0.22 (0.29) & 0.16(0.03) \\
 \hline
\end{tabular}
\end{center}

\clearpage
\section{BenchMarking Noise Aware Approaches}
\label{appendix:noiseawarebenchmark}
 \textbf{Overview \& Experimental Setup.} We now assess whether training models with noise-aware algorithmic interventions (e.g. robust loss functions~\citep{ma2020normalized, ghosh2017robust}) results in models whose disparity metrics have reduced sensitivity to label error in the training set. We test this hypothesis on a modified Cifar-10 dataset following the setting of ~\citet{hall2022systematic}. Specifically, the Cifar-10 dataset is modified to a binary classification setting along with group labels by inverting a subset of each class's examples. Given a specified parameter $\epsilon \in [0, 1/2]$, a $\frac{1}{2} - \epsilon$ of the negative class is inverted, while a $\frac{1}{2} + \epsilon$ of the positive class is inverted leading to $2\epsilon$ fraction of one group of samples and $1-2\epsilon$ of the other group. In all experiments we set $\epsilon=0.15$ for a 30 percent minority group membership. We replicate the label flipping experiment on the task with a Resnet-18 model. We test the MEIDTM~\citep{cheng2022instance}, DivideMix~\citep{li2020dividemix}, and a robust loss approach~\citep{ghosh2017robust}. We train the Resnet-18 with default parameters from~\citet{hall2022systematic}.

 \textbf{Results.}  At a high level, for the majority group, we find that model accuracy and downstream disparity metrics remain resilient to low rates of label error (below 25 percent). At high rates (>30 percent label error), we start to see declines in these performance metrics. However, for the minority group (30 percent of the dataset), we observe that the disparity metrics show consistent decline as label error is injected in these groups. This finding suggests that noise-aware methods show disparate performance in their ability to confer robustness to label error depending on data group sizes. A similar observation has also been made for other algorithmic interventions like Pruning~\citep{tran2022pruning, hooker2019compressed}, differential privacy~\citep{bagdasaryan2019differential}, selective Classification~\citep{jones2020selective} and adversarial training~\citep{xu2021robust}.

\clearpage
\section{Theoretical Insights for Automatic Relabeling}
\label{appendix:theory}
In this Section, we will prove Theorem~\ref{thm:relabeledcalibration}. To do this, we will appeal to two Lemmas from recent work. First, we restate the theorem here: 

\begin{theorem}
Given a $\kappa$-strongly convex loss function $\ell(., .)$, with $\kappa > 0$, a training dataset, $\mathcal{D}$, where $A$ indexes the data groups, and a model, $\hat{\theta} : x_i \rightarrow y_i$, optimized via $\ell(., .)$ that maps inputs to labels. Let $\mathcal{Q}$ be a set of test examples all belonging to group $A=a$, where $\mathrm{ECal}_{\mathcal{Q}}(\hat{\theta})$ is the expected calibration error of $\hat{\theta}$ on the set $\mathcal{Q}$. In addition, let $\mathcal{D}_{A=a}$ be the set of problematic training examples, belonging to group $a$, prioritized based on influence, i.e., $\mathcal{I}_{\mathrm{pert}, \mathrm{calib}, y^i}(x^i_a, \mathcal{Q}) > 0$. We term a model trained on a different training set ($\mathcal{D}_{+}$) where the problematic examples have been relabeled to be $\hat{\theta}_R$. Analogously, the expected calibration error of this new model on the set $\mathcal{Q}$ is $\mathrm{ECal}_{\mathcal{Q}}(\hat{\theta}_R)$. We have that:
\begin{center}
$\mathrm{ECal}_{\mathcal{Q}}(\hat{\theta}_R) \leq \mathrm{ECal}_{\mathcal{Q}}(\hat{\theta})$.
\end{center}
\label{thm:relabeledcalibrationappendix}
\end{theorem}

First, we will state Lemma that bounds the calibration of a classifier in terms of its excess risk. ~\citet{liu2019implicit} study unconstrained ERM and the associated calibration of the resulting model. They give a bound on the calibration of a model obtained via ERM in terms of its excess risk.

\begin{lemma}[Theorem 2.3 from~\citet{liu2019implicit}]
Given a $\kappa$-strongly convex loss function $\ell(., .)$, with $\kappa > 0$, let $\ell^\ast$ be the population risk of the Bayes classifier, then $\mathrm{ECal}(\hat{\theta})$, the expected calibration error of a classifier, $\hat{\theta}$, can be bounded as:  
\begin{center}
$\mathrm{ECal}(\hat{\theta}) \leq 2 \sqrt{\frac{\ell(\hat{\theta}) - \ell^\ast}{\kappa}}$.
\end{center}
\label{calibrationboundlemma}
\end{lemma}

We refer readers to~\citet{liu2019implicit} for the proof. In the Lemma, we are able to bound the group expected calibration error for an ERM solution relative to that of the best classifier in the model class. The strategy we will take to prove Theorem~\ref{thm:relabeledcalibration} will be to relate the bound from the Lemma~\ref{calibrationboundlemma} for two different classifiers. We will compare the bound for a classifier trained on a dataset without relabelled training samples to one trained on a data that has been relabeled. In particular, the relabeling will be according to the influence of a change in a training point's label on the expected loss on a particular test set.

We now define a few quantities. We let: $\mathcal{Q}$ represent the collection of examples in the test set. For now, we will restrict $\mathcal{Q}$ to consist of examples for a single group $a \in A$. Here $A \in [1, \ldots, \vert A \vert]$ indexes the groups in the training set, $\mathcal{D}$. We will let a model trained on the original training set (no relabelling) to be:  $\hat{\theta}$. 

We recall that $\mathcal{I}_{\mathrm{pert}, \mathrm{calib}, y^i}(x^i_a, \mathcal{Q})$, defined from the main text, estimates the influence of changing the label of the training sample $x^i_a$ by an infinitesimal amount. Consequently, based on the relabelling scheme that we have defined in Section~\ref{correctlabelerror}, we can define a subset of the training set: $\mathcal{D}_{A=a}$, which would be examples for which $\mathcal{I}_{\mathrm{pert}, \mathrm{calib}, y^i}(x^i_a, \mathcal{Q}) > 0$. These are the problematic examples that need to be relabeled. The set of examples that corresponds to the relabeled version is $\mathcal{D}_{+}$. Since we are primarily focused on a binary classification task, the relabeled version is simply a bit flip of the original labels.

\begin{lemma}[Theorem 1 from~\citet{kong2021resolving}]
Given two binary classifiers: $\hat{\theta}$, with loss $\ell(\hat{\theta}, \mathcal{Q})$, and  $\hat{\theta}_R$ with loss  $\ell(\hat{\theta}_R, \mathcal{Q})$ on the test set $\mathcal{Q}$ respectively, then: 
\begin{center}
$\ell(\hat{\theta}_R, \mathcal{Q}) \leq \ell(\hat{\theta}, \mathcal{Q}) \leq 0$.
\end{center}
\label{lossrelabellemma}
\end{lemma}

Lemma~\ref{lossrelabellemma} indicates that the loss of the model trained on the related dataset is lower or equal to the loss of the model trained on the dataset where a problematic data points, according to influence, have been relabeled.

To conclude the prove of the Theorem, we will combine insights from these two previously stated lemmas. First, we have that the bound on the expected calibration error of $\hat{\theta}$ according to Lemma~\ref{calibrationboundlemma} is: 
$$ \mathrm{ECal}(\hat{\theta}) \leq 2 \sqrt{\frac{\ell(\hat{\theta}) - \ell^\ast}{\kappa}}.$$ 

Similarly, the bound on the expected calibration error of $\hat{\theta}_R$ according to Lemma~\ref{calibrationboundlemma} is:
$$ \mathrm{ECal}(\hat{\theta}_R) \leq 2 \sqrt{\frac{\ell(\hat{\theta}_R) - \ell^\ast}{\kappa}}.$$

From Lemma~\ref{lossrelabellemma}, we know that $\ell(\hat{\theta}_R, \mathcal{Q}) \leq \ell(\hat{\theta}, \mathcal{Q}) \leq 0$, consequently, we can conclude that: $\mathrm{ECal}_{\mathcal{Q}}(\hat{\theta}_R) \leq \mathrm{ECal}_{\mathcal{Q}}(\hat{\theta})$.

\section{Longer Conclusion}
\label{appendix:conclusion}

 In this work, we study the effect of label error, both in the training and test set, on a model's disparity metric. Fairness audits are typically used to help surface disparate performance across groups in the data. However, if either the test data used to perform a fairness assessment or the training data used to fit the model contains label error, then it is unclear whether the conclusions of such a fairness assessment are reliable. Given the widespread presence of label error in the ML labeling pipeline, understanding the downstream effects of label error on a model's disparity metric is important. In this paper, we sought to address two key questions: \textit{1) What is the impact of label error on a model's group disparity metric, especially for smaller groups in the data;} and \textit{2) How can a practitioner identify training samples whose labels would also lead to a significant change in the test disparity metric of interest?}

To address the first question, we conducted empirical sensitivity tests. First, we flip the labels in the test set, to simulate label error, then measure the corresponding change to the disparity metric. Second, we fix a test set, but instead iteratively simulate varying levels of label error in the training set. We then measure the change in disparity metric, e.g., group calibration error for models trained on these `contaminated' datasets compared to a model derived from data without label error. 

We find that disparity metrics are, indeed, sensitive to test and training time label error. In addition, for the same level of label error, the percentage change in group calibration error for the minority group is on average 1.5 times larger than the change for the majority group. A possible explanation for the training-time results is that label error affects the ability of the model to easily learn patterns for the minority group even at lower fractions of label error in the training data. 

Second, we present an approach for estimating the `influence' of perturbing a training point's label on a disparity metric of interest. We empirically assess the proposed approach on a variety of datasets and find a 10-40\% improvement, compared to alternative approaches, in identifying training inputs that improve a model's disparity metric.

Our findings come with certain limitations. In this work, we focused on the influence of label error on disparity metrics. However, other components of the ML pipeline can also impact downstream model performance. The proposed empirical tests simulate the impact of label error; however, it might be the case that real-world label error is less pernicious to model learning dynamics than the synthetic flipping results suggest. Ultimately, we see our work as helping to provide insight and as an additional tool for practitioners seeking to address the challenge of label error particularly in relation to a disparity metric of interest.

\end{document}